\documentclass[runningheads]{llncs}

\usepackage[]{eccv}
\usepackage{eccvabbrv}

\usepackage{graphicx}
\usepackage{booktabs}
\usepackage[normalem]{ulem}
\useunder{\uline}{\ul}{}
\usepackage[accsupp]{axessibility}  % Improves PDF readability for those with disabilities.

\usepackage[pagebackref,breaklinks,colorlinks,citecolor=eccvblue]{hyperref}
\usepackage{orcidlink}
\usepackage{multirow}
\usepackage[graphicx]{realboxes}

\usepackage[inkscapelatex=false]{svg}

\usepackage{enumitem}
\usepackage{amsmath}
\usepackage{hyperref}

\begin{document}

\title{Zero-Reference Lighting Estimation Diffusion Model for Low-Light Image Enhancement} 

\titlerunning{Zero-LED}

\author{Jinhong He \and
Minglong Xue\inst{\thanks{Corresponding author}} \and
Aoxiang Ning \and\\
Chengyun Song}
% Senming Zhong\inst{2}}}

\institute{Chongqing University of Technology, China \\
% Guangdong Polytechnic Normal University, China\\
\email{hejh@stu.cqut.edu.cn}, \email{xueml@cqut.edu.cn}, \email{ningax@stu.cqut.edu.cn}\\
\email{scyer123@163.com}
}

\maketitle

\begin{abstract}
Diffusion model-based low-light image enhancement methods rely heavily on paired training data, leading to limited extensive application. Meanwhile, existing unsupervised methods lack effective bridging capabilities for unknown degradation. To address these limitations, we propose a novel zero-reference lighting estimation diffusion model for low-light image enhancement called Zero-LED. It utilizes the stable convergence ability of diffusion models to bridge the gap between low-light domains and real normal-light domains and successfully alleviates the dependence on pairwise training data via zero-reference learning. Specifically, we first design the initial optimization network to preprocess the input image and implement bidirectional constraints between the diffusion model and the initial optimization network through multiple objective functions. Subsequently, the degradation factors of the real-world scene are optimized iteratively to achieve effective light enhancement. In addition, we explore a frequency-domain based and semantically guided appearance reconstruction module that encourages feature alignment of the recovered image at a fine-grained level and satisfies subjective expectations. Finally, extensive experiments demonstrate the superiority of our approach to other state-of-the-art methods and more significant generalization capabilities. We will open the source code upon acceptance of the paper.
\end{abstract}

\section{Introduction}
\label{sec:intro}
Low light enhancement aims to enhance the quality and brightness of under-illuminated images. Due to the complex lighting conditions in the real world, relevant information in captured images is often lost through appropriate or significant masking. This poses a challenge to human visual perception and impedes the development and deployment of various downstream tasks (e.g., Autonomous Driving \cite{li2021deep}, Target Detection \cite{liang2021recurrent}, Text Detection \cite{xue2020arbitrarily}). Low-light image enhancement techniques have been significantly developed to address these challenges recently, with many related algorithms proposed. These techniques can be broadly categorized into traditional model-based approaches and data-driven deep learning-based approaches. The former primarily involves constructing physical models through methods such as histogram equalization \cite{pisano1998contrast} and Retinex theory \cite{land1971lightness}, which are then processed through manual a priori optimization of model parameters and the information inherent in the image itself \cite{park2022histogram,fu2016weighted,lin2014multi,li2018structure}. The effectiveness of these traditional methods relies heavily on the accuracy of the manual prior assumptions. However, lighting conditions are inherently ill-defined in the real world, leading to difficulties elucidating various low-light factors.

\begin{figure}[tb]
        \centering
        \includegraphics[height=0.4\textwidth,width=\textwidth]{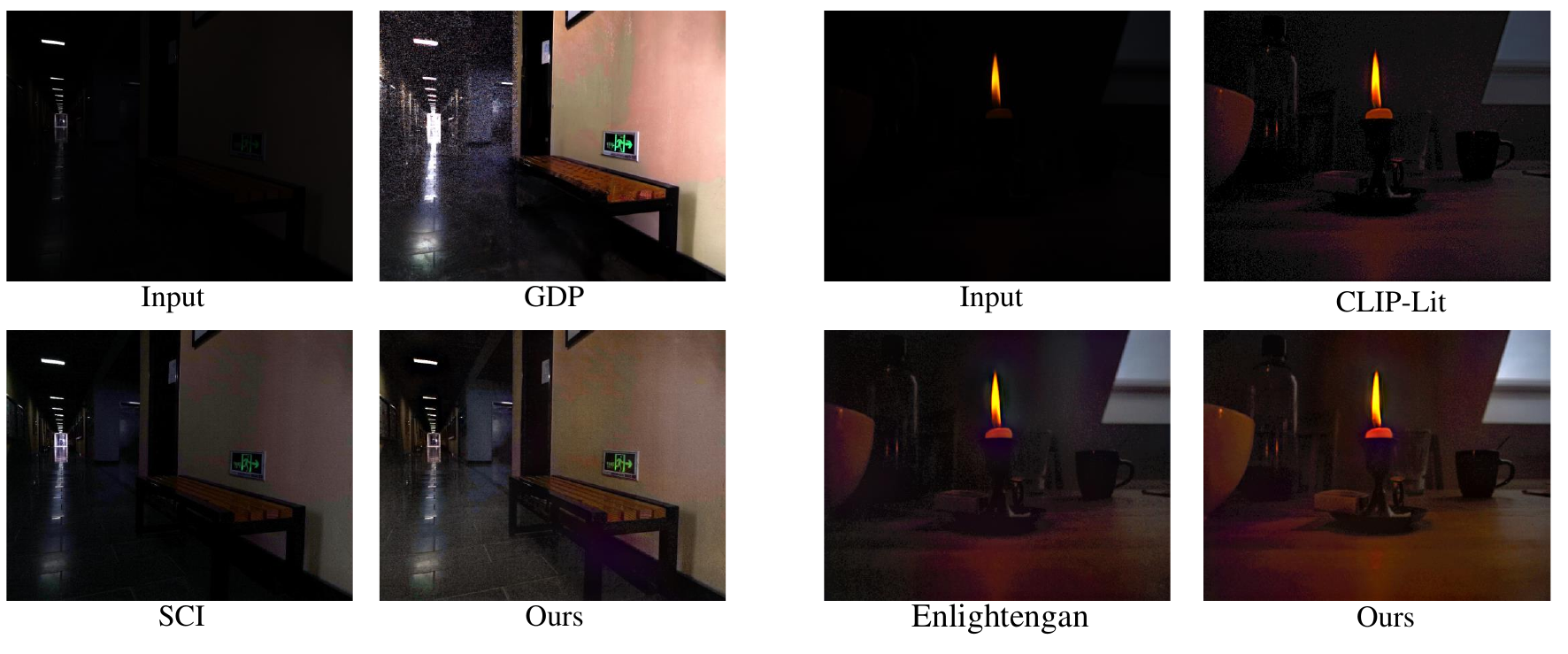}  
         \vspace{-2em}
        \caption{Comparison between state-of-the-art unsupervised methods and our method. It can be seen that these comparison methods appear to suffer from excessive noise, color distortion, and visual quality degradation.}
        \label{fig:1}
\end{figure}

With the development of deep learning, the quest to extract a priori knowledge from massive datasets has given rise to numerous data-driven network learning methods \cite{xu2022snr,sun2015learning,jiang2023low}. Supervised learning-based methods utilize paired low-light/high-light image data to establish the corresponding association mappings between low-light and clear images for learning directly, which is more robust and effective than traditional methods. Despite these advances, there are still significant challenges in constructing paired training data, which has led to the exploration of low-light enhancement methods based on unsupervised learning \cite{guo2020zero,liu2021retinex,jiang2021enlightengan,liang2023iterative}. Essentially, they all work by bridging the mapping between the input and output domains to obtain clear highlighted images. However, the existing unsupervised methods make it challenging to generate a reconstruction of the content structure due to the lack of effective supervision. They can also not effectively generate and optimize the unknown degradation factors of realistic complex scenes, which often cause excessive noise and artifacts to the extent that it is difficult to obtain satisfactory visual restoration, as shown in Fig. \ref{fig:1}. In particular, numerous low-light image enhancement algorithms based on the diffusion model \cite{jiang2023low,fei2023generative} have been proposed recently and achieved strong performance results. However, due to the stochastic nature of the diffusion inference process and the dependence on supervisory constraints, most of these algorithms are based on supervised training with paired datasets or optimizing the network using a priori knowledge, which makes it challenging to achieve genuinely effective unsupervised diffusion training and widely deployed in real-world.

Therefore, to alleviate the above problems, we explore a bidirectionally constrained unsupervised diffusion training approach to address the lack of generative power during unsupervised training and the unknown degradation of realistic low-light scenes. Specifically, we propose a light estimation-based diffusion model for zero-reference low-light image enhancement called Zero-LED. A pluggable initial optimization network is constructed via a deep network \cite{guo2020zero} for the preprocessing of diffusion, which is passed as a structural constraint to the diffusion process to mitigate the stochastic nature of diffusion. Subsequently, the light estimation of the inference process is optimized by an objective function, and this is passed in reverse to optimize the initial optimization network to achieve iterative bidirectional supervisory optimization. In addition, to reduce the consumption of computational resources, we transfer the diffusion process to be carried out on the wavelet domain by wavelet transform. we design a text- and frequency-domain based Appearance Reconstruction Module (ARM), which guides the inference output of diffusion through different modalities and combines the efficient capture of detailed content in the frequency-domain space to achieve metrics-favorable and perceptually oriented enhancement effects. Benefiting from these, our approach has a more significant generalization capability to the real world, and extensive experiments on publicly available datasets demonstrate the superiority of our approach over other state-of-the-art unsupervised methods. Overall, our contributions are as follows:
\begin{itemize}[noitemsep,topsep=10pt]
\item We propose a bidirectional optimised unsupervised training method that effectively implements a low-light image enhancement diffusion model without relying on reference images, thus reducing the dependence on paired training data. The model has enhanced generative capabilities and seamlessly bridges the normal and low-light domains, introducing new perspectives for future research.
% We have designed a bidirectional optimized unsupervised training method to effectively implement a low-light image-enhanced diffusion model without relying on reference images, thereby reducing dependence on paired training data. This model exhibits heightened generative capabilities and seamlessly bridges the gap between normal-light and low-light domains, introducing fresh perspectives for future research.
\item We design a semantic and frequency domain-based appearance reconstruction module. It utilizes different modalities and multiple frequency domain spaces to constrain the stochastic nature of the diffusion inference process and efficiently reconstructs images for better perceptual results.
\item Extensive experiments on real-world based datasets have demonstrated the superiority of our method over other state-of-the-art methods, as well as more significant generalization capabilities.
\end{itemize}

\section{Related Work}

\subsection{Low-light Image Enhancement}
Early techniques for enhancing low-light images primarily focused on employing a model-based approach to optimize the parametric information within the image itself for processing. The Retinex theory \cite{land1971lightness} achieves the desired reflectance map (standard image) by eliminating the low-light input's illumination. Wang et al. \cite{wang1999image} developed an adaptive histogram equalization method, and Pisano et al. \cite{pisano1998contrast} proposed contrast-adaptive histogram equalization methods. However, these model-based approaches often depend on manual crafting of many a priori assumptions and struggle to adapt to the complex environmental conditions of natural scenes.

Therefore, along with the proven benefits of deep learning in numerous low-level vision tasks, researchers have also focused their attention on low-light image enhancement algorithms \cite{wei2018deep,li2023embedding,xu2022snr,wu2022uretinex,zamir2020learning,zamir2022restormer}  that leverage a data-driven approach for efficient generalization through deep learning. For example, Chen et al. \cite{chen2018deep} curated a dataset containing different exposure levels for nighttime imaging challenges. Wei et al. \cite{wei2018deep} designed an end-to-end trainable RetinexNet. However, due to the challenges of acquiring paired low-light image datasets, Jiang \cite{jiang2021enlightengan} utilized generative adversarial Networks (GANs) as the main framework, pioneering unpaired images for training. Similarly, Guo et al. \cite{guo2020zero} elaborated a pixel-level curve estimation convolutional neural network by iterative derivation to establish a reference-free training paradigm. Liang et al. \cite{liang2021recurrent} designed a Retinex architecture-based search unfolding technique. Yang et al. \cite{yang2023implicit} employed neural representation to normalize the degradation to alleviate the enhancement difficulty. Despite the achievements of these unsupervised methods, their generalization to real-world scenarios is still limited. For this reason, we adapt to various complex environmental conditions by diffusing the generative power of the model.

\subsection{Diffusion Model for Image Restoration}
Recently, diffusion models \cite{song2019generative,song2020score} have garnered significant acclaim within image generation by leveraging parametric Markov chains to optimize the lower variational bounds of the likelihood function. This enables them to yield more precise target distributions than alternative generative models, such as GANs. Concurrently, to amplify the generative prowess of algorithms in image restoration, many researchers have embarked on developing various restoration endeavours grounded in diffusion models. Saharia et al. \cite{saharia2022image} adopt a direct cascading approach, integrating low-resolution measurements and latent codes as inputs to train conditional diffusion models for restoration. Jiang et al. \cite{jiang2023low} advances a diffusion model rooted in wavelet transform tailored for enhancing images captured in low-light environments, achieving content stabilization through forward diffusion and denoising processes during training. WeatherDiff \cite{ozdenizci2023restoring} introduces a block-based diffusion model aimed at recuperating images taken in adverse weather conditions, employing guidance across overlapping blocks during the inference stage. Additionally, Fei et al. \cite{fei2023generative} utilize the a priori knowledge embedded in a pre-trained diffusion model to effectively address any linear inverse problem.

Although diffusion models have achieved satisfactory visual restoration, due to the uncontrollable nature of diffusion, these algorithms are almost always based on supervised training on paired datasets or network optimization using the a priori knowledge of pre-trained diffusion models. It is still a great challenge to realize unsupervised training. In this paper, we propose a bi-directionally constrained unsupervised diffusion training approach to achieve robust zero-reference trained diffusion models for the first time, as well as more significant generalization ability and effective low-light image enhancement.

% \subsection{Multi-modal in Vision}
% In recent years, there has been a remarkable surge of interest among researchers in the realm of efficient cross-modal learning \cite{zang2022open,kuo2022f,zhou2022extract}. This trend has paved new pathways in computer vision and led to the emergence of numerous visual language models, culminating in significant achievements. As exemplified by CLIP, Radford et al. \cite{radford2021learning} advocated extracting prior knowledge from vast image-text datasets to construct visual language models. Leveraging a colossal knowledge base of 400 million image-text pairs, CLIP demonstrates the capacity for efficient task migration through zero-shot training. 

% Recent studies \cite{liang2023iterative,wang2023exploring} have shown that rich visual-linguistic a priori can be used to assess image quality and abstraction perception. Inspired by these studies, we plan to employ a priori knowledge from pre-trained visual language models to guide the diffusion inference process for low-light image enhancement. The accuracy of the semantic information is used to reduce the randomness of diffusion and achieve better performance.

\begin{figure*}[t]
        \centering
        \includegraphics[height=0.66\textwidth,width=1.01\textwidth]{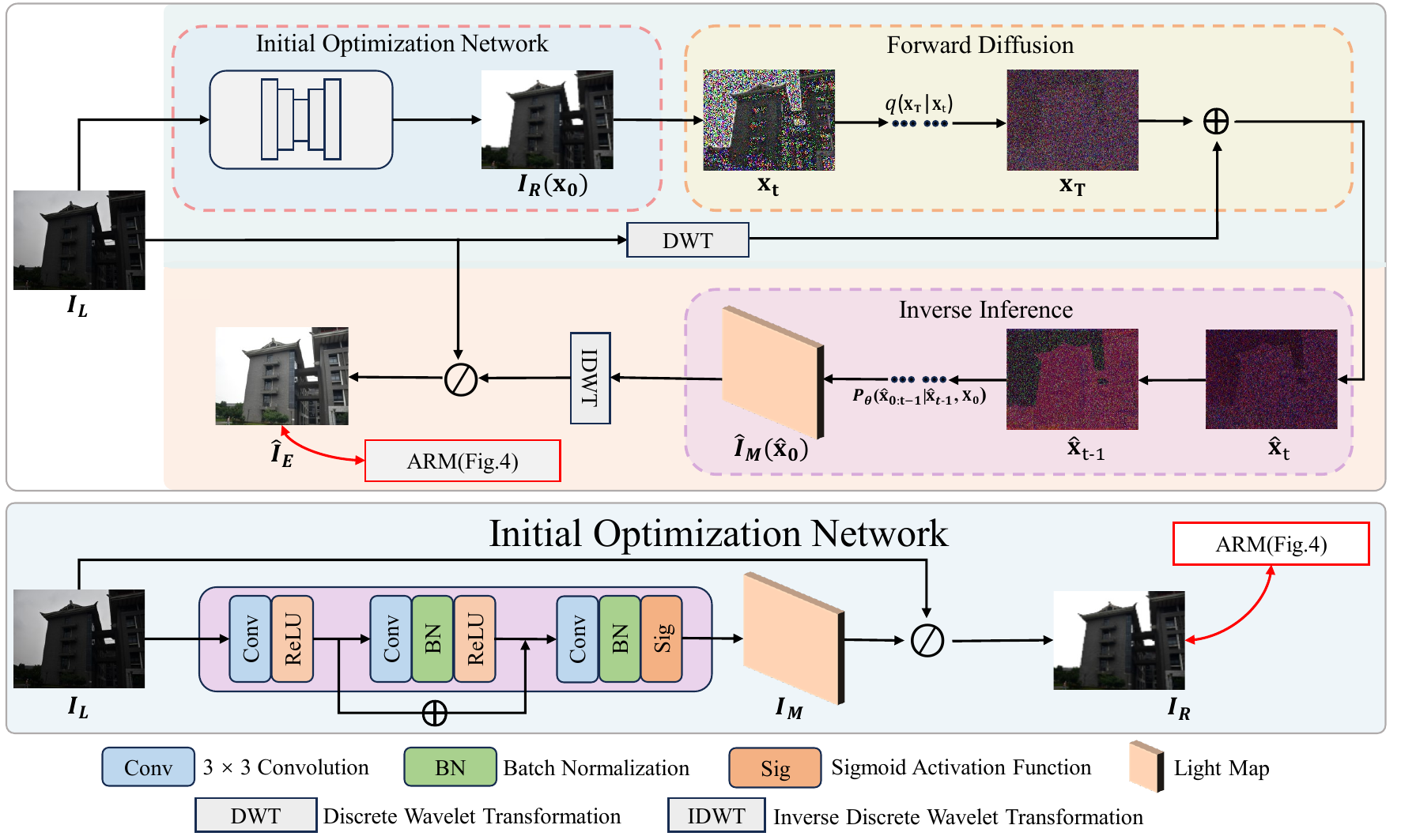}
        \caption{The overall framework of our proposed Zero-LED is illustrated. It proposes a Bidirectional optimization approach combining a deep neural network and a diffusion model for training without reference images. The initial optimization network provides the structural image and preliminary optimization of unknown degradation factors for the diffusion process. The inference process further bridges the gap between degraded and normal light and is optimized by an objective function in both directions. Meanwhile, We effectively reduce the consumption of computational resources through wavelet transform. The bottom part shows in detail the pluggable initial optimization network we designed.}
        \label{fig:2}
\end{figure*}
% \vspace{-1em}

\section{Methodology}
The overall framework of our proposed Zero-LED is illustrated in Fig. \ref{fig:2}. Leveraging the generative capacity of the diffusion model, the proposed method achieves notable enhancements in image quality. By developing a bidirectional optimization training method, we establish a diffusion model based on zero-reference images, thereby reducing reliance on training data and enhancing generalizability to real-world contexts. Furthermore, to minimize computational resource consumption and enhance efficiency \cite{jiang2023low}, we transition the diffusion inference process to the wavelet low-frequency domain via wavelet transformation. In this section, we provide a detailed exposition of the underlying principles of the traditional conditional diffusion model and the crux of our proposed methodology, namely, the bidirectional optimization-based zero-reference diffusion model. Lastly, we introduce the Appearance Reconstruction Module (ARM), grounded in text and frequency domains, as a meticulously crafted component adept at guiding the reconstruction of image content structure and the overarching enhancement of quality.

\subsection{Conditional Diffusion Models}
Diffusion models \cite{ho2020denoising,song2020denoising} to train Markov chains by variational inference. It converts complex data into completely random data by adding noise and gradually predicts the noise to recover the expected clean image. Consequently, it usually includes the forward diffusion process and reverse inference process.

{\bf Forward Diffusion Process.} The forward diffusion process primarily relies on incrementally introducing Gaussian noise with a fixed variance $\{\beta_t\in(0, I)\}_{t=1}^T$ into the input distribution ${x}_0$ until the T time steps approximate purely noisy data. This process can be expressed as:
\vspace{-6pt}
\begin{equation}
q(x_1,\cdots,x_T  | x_0) =\prod_{t=1}^{T}q(x_t  |x_{t-1}),
\tag{1}
\end{equation}

\begin{equation}
q(x_t|x_{t-1})=N(x_t;\sqrt{1-\beta_t}x_{t-1},\beta_tI),
\tag{2}
\end{equation}
where $x_t$ and $\beta_t$ are the corrupted noise data and the predefined variance at time step $t$. Respectively, $N$ denotes a Gaussian distribution. Furthermore, each time step $x_t$ of the forward diffusion process can be obtained directly by calculating the following equation, computed from $x_0$:

\begin{equation}
x_t=\sqrt{\overline{\alpha}_t}x_0+\sqrt{1-\overline{\alpha}_t}\epsilon,\ \  \epsilon \sim N(0, I),
\tag{3}
\end{equation}
where $\alpha_t=1-\beta_t$, $\overline{\alpha}_t$=$\prod_{i=1}^{t}\alpha_i$.

{\bf Reverse Inference Process.} The reverse inference process aims to restore the original data from the generated Gaussian noise. In contrast to the forward diffusion process, where the data distribution at each time step $t$ can be directly computed using a formula, the reverse process relies exclusively on iteration to eliminate the predicted noise and restore the data until the randomly sampled noise $x_T\sim N(0, I)$ into the clean data $\hat{x}_0$. Formulated as:
\begin{equation}
p_\theta(\hat{x}_0,\cdots,\hat{x}_{T-1}  | x_T)=\prod_{t=1}^{T} p_\theta(\hat{x}_{t-1}|\hat{x}_t),
\tag{4}
\end{equation}

\begin{equation}
p_\theta(\hat{x}_{t-1}|\hat{x}_t)=N(\hat{x}_{t-1};\mu _\theta(\hat{x}_t,t),\sigma ^2_tI),
\tag{5}
\end{equation}
where $\mu _\theta$ is the diffusion model noise predictor, which is mainly optimized by the editing and data synthesis functions and used as a way to learn the conditional denoising process, as follows:
\begin{equation}
\mu _\theta=\frac{1}{\sqrt{\alpha_t}}(x_t-\frac{\beta_t}{\sqrt{1-\overline{\alpha}_t}}\epsilon_\theta(x_t,t)),
\tag{6}
\end{equation}
where $\epsilon_\theta$ is a function approximator intended to predict $\epsilon$ from $x_t$.

\subsection{Diffusion Models for Zero-reference Learning}

% While traditional conditional diffusion models exhibit significant performance, the substantial demands on computational resources and paired datasets remain formidable challenges for the generalization capability and practical deployment of diffusion models in image restoration tasks. To address these issues, we introduce a zero-reference learning approach for diffusion models, consisting of the following components:
Existing conditional diffusion models have achieved significant performance, but the substantial demands on computational resources and paired datasets hinder their practical deployment in image restoration tasks. To address these challenges, we propose a zero-reference learning approach for diffusion models, which includes the following components:

{\bf Initial Optimization Network.} To consider the generative diversity of the diffusion model, we use an initial optimization network as the a priori network for diffusion training. Specifically, we do this by preprocessing the input image and passing it into the diffusion process. Compared with the input image, the preprocessed structural image $I_R$ has clearer image content. The generation of chaotic content can be constrained more precisely in the inference process. In addition, the initial optimization network provides an initial degradation domain calibration for the diffusion model and combines the objective function to form an iterative optimization with the diffusion model.

% For the consideration of generative diversity of the diffusion model, we use the initial optimization network as the a priori network for diffusion training. The random content of the diffusion model is constrained by preprocessing the input images and passing them into the diffusion process. In addition, the Initial Optimization Network can provide an initial degradation domain calibration for the diffusion model, which forms an iterative optimization jointly with the diffusion model during training. As shown in Fig. (####), the structural images preprocessed by the Initial Optimization Network provide more image information for the diffusion recovery process, and the optimization of the degradation parameters is more significant compared to the input low-luminance images.

Based on a kind of consensus \cite{ma2022toward,guo2020zero} that there is a link between illumination and low-light images, as well as the consideration of model complexity. Therefore, we plan to decompose the low light image $I_L$ using a lightweight deep neural network to obtain its illumination component $I_M \in R^{H\times W\times C}$. Additionally, based on the Retinex theory (i.e. $I_L$ = $I_R$ $\otimes$ $I_M$), we can preliminarily obtain the structural image $I_R$.

{\bf Diffusion-based Degradation Model.} Inspired by Jiang et al. \cite{jiang2023low}, we utilize the wavelet transform to process the input image. The use of the discrete wavelet transforms $( DWT(\cdot))$ enables the extraction of the low-frequency space of the low-light image, thereby reducing the computational resources required by the diffusion model. The input image is decomposed into its low-frequency information $(L)$ and high-frequency information $(H)$ through the $ DWT(\cdot)$. Furthermore, the high-frequency space $(H)$ produced by the wavelet transform includes three subbands in the vertical, horizontal, and diagonal directions. As a result, the low-frequency information that has been decomposed $(L)$ can be considered a version of the image that has been reduced by a quarter dimension, significantly reducing the demand for computer resources. 

In the real world, images are often subject to unknown degradation factors in complex, dimly lit environments, and thus, low-light image enhancement is often seen as a task to construct unknown degradation models. We further simulate complex degradation based on the structured images $I_R$ provided by the initial optimization network and combined with the generative power of the diffusion model. The diffusion model is also used to generate fine-tun illumination masks to achieve significant enhancement effects. This can be described as:

\begin{equation}
\hat{I}_E = I_L \oslash \mathcal{D}(I_R,I_L),
\tag{7}
\end{equation}
where $\mathcal{D}$ is a diffusion-based degenerate model. Specifically, the inference process Eq. {\color[HTML]{FF0000}4} $p_\theta(\hat{x}_0,\cdots,\hat{x}_{T-1}  | x_T)$ will be carried out under the structural image $I_R$ and low light image $I_L$. Our goal is to learn the degradation parameters of the image from the denoising process $p_\theta(\hat{x}_{0: T}|(I_R, I_L))$ while guaranteeing high fidelity of the sampling results to its generated content. Additionally, we will optimize the noise predictor to fit the illumination component $\hat{I}_M$ and minimize the L2 distance between $\hat{I}_M$ and $I_M$ to refine its degeneracy parameters and optimize the initial optimization network. Therefore, the objective function of the optimized diffusion model can be expressed as follows:

\begin{equation}
\mathcal{L}_{diff}=E_{t\sim[1,T]}E_{x_0\sim p(x_0)}E_{\epsilon_t\sim N(0,I)}\parallel \epsilon_t-\epsilon_\theta(x_t,t)\parallel^2+\ \vert\vert\hat{I}_M-{I}_M \vert\vert^2.
\tag{8}
\end{equation}

In addition, smoothing loss with a spatially varying number of paradigms \cite{ma2022toward,fan2018image} is used to optimize the predicted illumination component, denoted as:

\begin{equation}
\mathcal{L}_{smooth}=\sum_{k=1}^{K} \sum_{n \in K(k)}\gamma_{k,n}(\parallel \hat{I}_{M(k)}-\hat{I}_{M(n)}\parallel_1 +\parallel I_{M(k)}-I_{M(n)}\parallel_1),
\tag{9}
\end{equation}
Where $K$ is the total number of pixels, $k$ is the kth pixel. $\gamma_{k,n}$ denotes the weight.

\begin{figure*}[t]
        \centering
        \includegraphics[height=0.63\textwidth,width=1.01\textwidth]{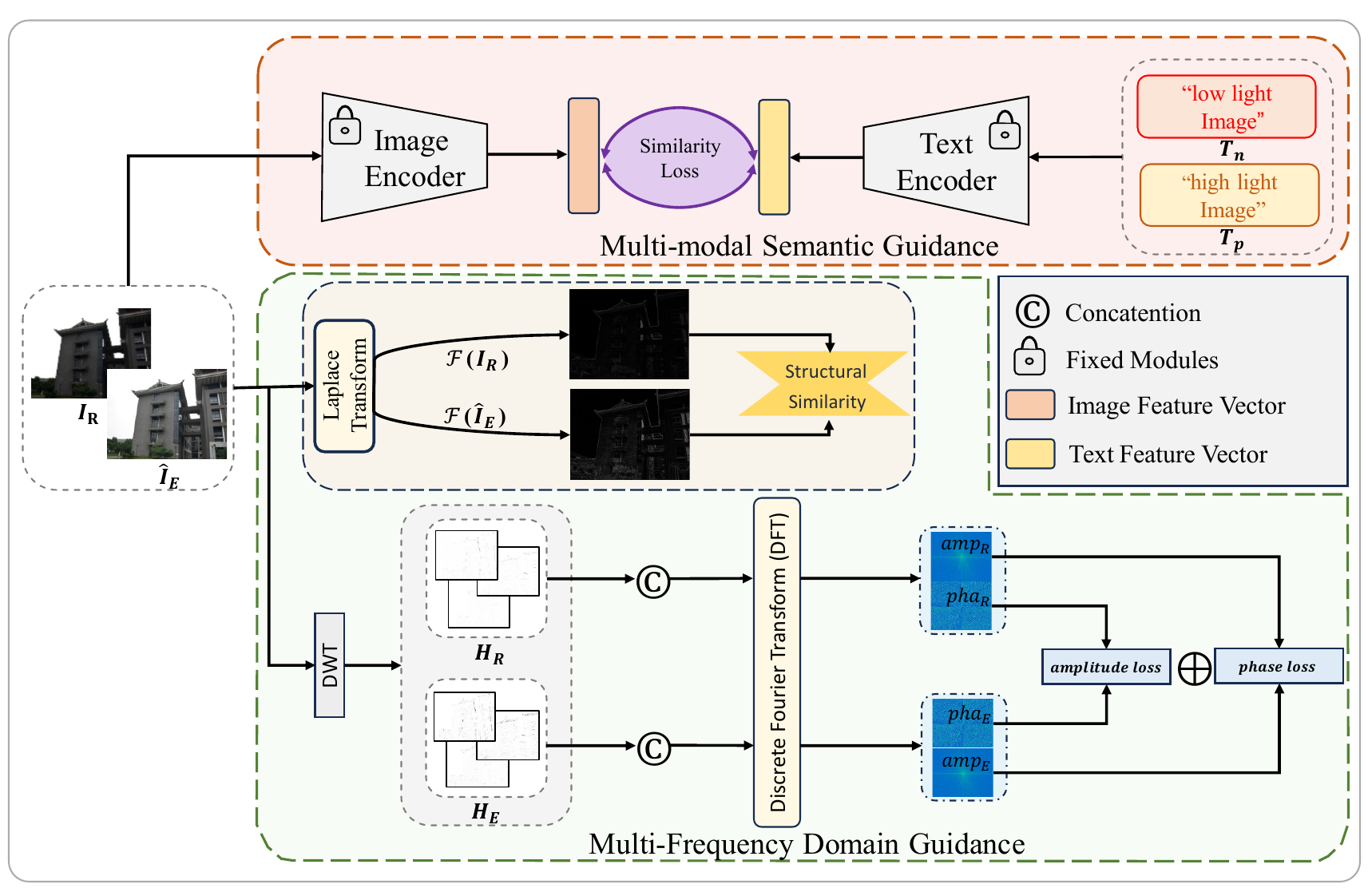}
        \caption{Framework diagram of our proposed appearance reconstruction module. multi-modal semantics focuses on guiding illumination enhancement and supervising the input of image and text features. Frequency-domain guidance focuses on supervising high-frequency details and constraining the generation of artifacts.}
        \label{fig:3}
    \end{figure*}

\subsection{Appearance Reconstruction Module}

Existing unsupervised methods hardly obtain significant detail features from low-light images for satisfactory restoration, while diffusion models are complex to perform effective content structure recovery in unsupervised training scenarios. Therefore, as shown in Fig. \ref{fig:3}, to achieve good reconstruction of the generated content and perceptually oriented enhancement, we propose a semantic and multi-frequency domain guided appearance reconstruction module based on semantic and multi-frequency domain guidance to obtain efficient appearance reconstruction.
% Therefore, as shown in Fig. \ref{fig:3}, for fine-grained feature recovery as well as content reconstruction, we propose a semantic-based and multiple frequency-domain-guided appearance reconstruction module to obtain efficient appearance reconstruction.

{\bf Multi-modal Semantic Guidance.} Recent studies have shown \cite{liang2023iterative,yang2023implicit} that multi-modal learning can lead to the effective enhancement of low-light images compared to a single modality. Therefore, we use a pre-trained CLIP model to acquire prior knowledge. Precisely, we extract the latent codes of the cue pairs by feeding predefined cue pairs containing positive prompts $T_p$ and negative prompts $T_n$ (as shown in Fig. \ref{fig:3}) to the text encoder ($\Phi _{text}$). Correspondingly, the latent codes of the prediction result $\hat{I}_E$ and the structure image $I_R$ are extracted by the image encoder ($\Phi _{image}$). We then measure the difference between image vectors and text vectors by computing the similarity loss between them in the CLIP latent space:

\begin{equation}
\mathcal{L}_{clip}=\sum_{I \in \{I_E,\hat{I}_E\}} \frac{e^{\cos(\Phi_{image} (I ),\Phi_{text}(T_n ))} }{\sum _{T\in\{T_p,T_n\}} e^{\cos(\Phi_{image} (I),\Phi_{text}(T))}}.
\tag{10}
\end{equation}

In addition, we present the hyper-parameters $\upsilon$ as the probability value of normal light to incentivize the prediction results further to achieve good illumination. Semantically consistent output is encouraged by minimizing the distance between the positive prompts probability and the hyper-parameters $\upsilon$:

\begin{equation}
\mathcal{L}_{prob}=\parallel cos(\Phi_{image}(\hat{I}_E ),\Phi_{text}(T_n ))-\upsilon||_1.
\tag{11}
\end{equation}

% Thus the total loss of textual guidance we can summarise as:
% \begin{equation}
% \mathcal{L}_{text}=\mathcal{L}_{prob}+\mathcal{L}_{clip}
% \tag{12}
% \end{equation}

{\bf Frequency Domain Guidance.} In this study, we combine the advantages of spatial information capture in multiple frequency domains. The spectrum is utilized to help diffusion models perform appearance reconstruction during unsupervised training, leading to metric-friendly and perception-oriented enhancements.
% In this study, we incorporate the advantages of multiple frequency domain spaces to effectively enhance the recovery of prediction results at the fine-grained level for metric-favorable and perception-oriented enhancement.

We first perform the Laplace transform $\mathcal{F}(\cdot )$ on the structural image $I_R$ to supervise the sampling results from the edge level. Simultaneously, we implemented constraints on the content of the image generation using SSIM content loss \cite{wang2004image} with an initial optimization network:
% Simultaneously, we utilize SSIM content loss \cite{wang2004image} to form a bi-directional constraint with the initial optimization network to achieve constraints on image content:

\begin{equation}
\mathcal{L}_{content}=(1-SSIM(\mathcal{F}(\hat{I}_E),\mathcal{F}(I_R)))+(1-SSIM(\hat{I}_E,I_R)).
\tag{12}
\end{equation}

In addition, we double-transform the content information of the image by combining the wavelet transform and the Fourier transforms to capture the deeply detailed features of the image and reconstruct the image using the frequency domain. Specifically, we take the discrete Fourier transform $\rm DFT(\cdot)$ the wavelet high-frequency domain of the predicted image $\hat{I}_E$ and the structural image $I_R$ to obtain their corresponding amplitude and phase (amp, pha):

\begin{equation}
amp_E,pha_E={ DFT}( DWT(\hat{I}_E)),
\tag{13}
\end{equation}
\begin{equation}
amp_R,pha_R={ DFT}( DWT(I_R)).
\tag{14}
\end{equation}

To encourage the recovery results to be feature-aligned at a fine-grained level and cross-validated with the optimization network. For this purpose, we use L1 loss to minimize the information differences between spectrograms:
\begin{equation}
\mathcal{L}_{spectral}=\vartheta _1 \parallel amp_E -amp_R \parallel_1 +\ \vartheta _2 \parallel pha_E- pha_R \parallel_1 ,
\tag{15}
\end{equation}
where $\vartheta_1$ and $\vartheta_2$ are weighting parameters for amplitude and phase losses. Thus, for multi-modal semantics and frequency domain-guided appearance reconstruction, the total loss can be summarised as:
\begin{equation}
\mathcal{L}_{rec}=\mathcal{L}_{content}+ \mathcal{L}_{spectral} + \varpi (\mathcal{L}_{prob} + \mathcal{L}_{clip}),
\tag{16}
\end{equation}
where $\varpi$ is the weight of the semantic guidance loss.

\subsection{Network Training}
Besides the objective functions used to optimise the diffusion model and appearance reconstruction, we also utilise two quality-enhancing losses and the MSE to improve the quality of the final output and extend the network learning capabilities.

{\bf Color Constancy Loss.} Based on the grey world color constancy hypothesis \cite{buchsbaum1980spatial}. We designed a color constancy loss to correct for potential color bias in the enhanced image and also established a relationship between the three adjustment channels:
\begin{equation}
\mathcal{L}_{col}=\sum_{\forall (m,n)\in \varrho  }(C^m-C_n)^2,  \varrho=\{(R,G) ,(R,B), (G,B)\},
\tag{17}
\end{equation}
where $C^m$ is the average intensity value of $m$ channel in the recovered image, $(m,n)$ represents a pair of channels.

{\bf Spatial Consistency Loss.} The loss of spatial coherence constraints the differences in neighbouring regions between the input image and the enhanced image:
\begin{equation}
\mathcal{L}_{spa}=\frac{1}{N} \sum_{i=1}^{N}\sum_{j\in \Theta(i) }(\parallel E_i-E_j \parallel_1 - \parallel I_i-I_j \parallel_1)^2,
\tag{18}
\end{equation}
where $N$ is the number of local regions and $\Theta(i)$ is the four neighbouring regions (top, bottom, left, and right) centred on region $i$. We denote $E$ and $I$ as the average intensity values of the local regions in the enhanced version and the low-light image, respectively.

The total loss $\mathcal{L}_{total}$ is expressed by combing the diffusion objective function, the appearance reconstruction loss, and the Quality Enhancement Losses as:
\begin{equation}
\mathcal{L}_{total}=\mathcal{L}_{diff}+\omega \mathcal{L}_{smooth}+\mathcal{L}_{rec}+\mathcal{L}_{col}+\mathcal{L}_{spa},
\tag{19}
\end{equation}
where $\omega$ is the weight of the illumination smoothing loss.

\section{Experiments}
\subsection{Experimental Settings}

{\bf Implementation Details.} We implemented our method using Pytorch on two NVIDIA RTX 3090 GPUs. We set the total number of training iterations to $5\times10^4$, using the Adam optimizer with the initial learning rate set to $1\times10^{-4}$. The batch size and patch size were set to $4$ and $256\times256$, respectively. 

{\bf Benchmark Datasets.} Our network is trained and evaluated to validate the effectiveness of our method on the LSRW \cite{hai2023r2rnet} dataset, in which we randomly select 1000 low-light-normal-light image pairs for training and 50 image pairs for evaluation. Most low-light images were collected realistically by varying the exposure time and ISO and fixing other camera parameters. In addition, we extend several real-world benchmark datasets to evaluate the performance of our proposed network to increase persuasiveness. Examples include LOLv1 \cite{wei2018deep}, LIME \cite{guo2016lime} and Backlit300 \cite{liang2023iterative}. The dataset of LOLv1 contains 500 real-world low/normal light image pairs, of which 485 image pairs are used for training and 15 image pairs are used for evaluation. Also, to demonstrate the generalization to real-world degraded scenes, we evaluate the generalization ability of the proposed method in this paper by the benchmark dataset LIME and by selecting 30 degraded images in the Backlit300 dataset. Note that during the training process, we only use low-light images from the paired dataset and do not use normal-light images to demonstrate the superiority of our zero-reference method.

{\bf Metrics.} We propose to evaluate the performance of different algorithms using two full-reference distortion metrics, PNSR and SSIM \cite{wang2004image}, and two non-reference metrics, NIQE \cite{mittal2012making} and LOE \cite{wang2013naturalness}. Higher PSNR or SSIM implies more realistic restoration results, while lower NIQE or LOE indicates higher quality details, luminance, and hue.

% \usepackage{multirow}
\begin{table*}[]
\renewcommand\arraystretch{1.6}
\scalebox{0.71}{
\begin{tabular}{l|c|cccc|cccc|cc|cc}
\hline
                          &                             & \multicolumn{4}{c|}{LOLv1}                                                                                                 & \multicolumn{4}{c|}{LSRW}                                                                                                    & \multicolumn{2}{c|}{LIME}                                      & \multicolumn{2}{c}{Backlit300}                                 \\ \cline{3-14} 
\multirow{-2}{*}{Methods} & \multirow{-2}{*}{Reference} & PSNR↑                          & SSIM↑                         & NIQE↓                         & LOE↓                            & PSNR↑                          & SSIM↑                         & NIQE↓                         & LOE↓                            & NIQE↓                          & LOE↓                            & NIQE↓                          & LOE↓                            \\ \hline
Zero-DCE                 & CVPR'20                     & 14.861                        & 0.559                        & 11.985                       & 215.816                        & 15.801                        & 0.446                        & 11.832                       & 247.291                        & 11.942                        & 192.089                        & 16.026                        & {\color[HTML]{00009B} 165.325} \\
Zero-DCE++                & TPAMI'21                    & 14.682                        & 0.472                        & 10.646                       & 277.736                        & 15.791                        & 0.457 & 11.341                       & 241.348                        & {\color[HTML]{00009B} 11.376} & 296.654                        & 14.693                        & 285.720                        \\
Enlightengan              & TIP'21                      & {\color[HTML]{FF0000} 17.606} & {\color[HTML]{FF0000} 0.653} & {\color[HTML]{00009B} 9.996} & 365.561                        & {\color[HTML]{FF0000} 17.136} & {\color[HTML]{00009B} 0.460} & 11.937                       & 385.135                        & 14.585                        & 421.018                        & 15.058                        & 385.796                        \\
RUAS                      & CVPR'21                     & 16.405                        & 0.499                        & 10.725                       & 125.351                        & 14.271                        & 0.411                        & 11.081                       & 198.930                        & 12.413                        & 288.730                        & 14.486                        & 598.305                        \\
SCI                       & CVPR'22                     & 14.784                        & 0.521                        & 11.827                       & {\color[HTML]{FF0000} 101.113} & 15.241                        & 0.419                        & 10.774                       & 234.605                        & 12.379                        & 212.621                        & {\color[HTML]{00009B} 13.376} & 298.768                        \\
CLIP-Lit                  & ICCV'23                     & 12.394                        & 0.493                        & 12.187                       & 355.441                        & 13.483                        & 0.405                        & {\color[HTML]{00009B} 9.144} & 289.583                        & 12.239                        & 192.001                        & 16.633                        & 195.875                        \\
GDP                       & CVPR'23                     & 15.896                        & 0.542                        & 10.273                       & {\color[HTML]{00009B} 120.278} & 12.887                        & 0.362                        & 9.178                        & {\color[HTML]{FF0000} 75.884}  & 13.138                        & {\color[HTML]{FF0000} 78.929} & 13.693                        & {\color[HTML]{FF0000} 148.929} \\ \hline
Ours                      & -                           & {\color[HTML]{00009B} 16.632} & {\color[HTML]{00009B} 0.566} & {\color[HTML]{FF0000} 8.355} & 148.563                        & {\color[HTML]{00009B} 15.824} & {\color[HTML]{FF0000}0.461 }                       & {\color[HTML]{FF0000} 8.381} & {\color[HTML]{00009B} 175.355} & {\color[HTML]{FF0000} 10.843} & {\color[HTML]{00009B} 146.663} & {\color[HTML]{FF0000} 11.993} & 351.877                        \\ \hline
\end{tabular}
}
\caption{Quantitative evaluation of different unsupervised learning methods on four benchmark datasets. The best and second performance are marked in {\color[HTML]{FF0000}{red}} and {\color[HTML]{00009B}{blue}}, respectively.}
\label{tab:my-table}
\end{table*}

\vspace{-1em}
\begin{figure*}[h]
        \centering
        \includegraphics[height=0.3\textwidth,width=\textwidth]{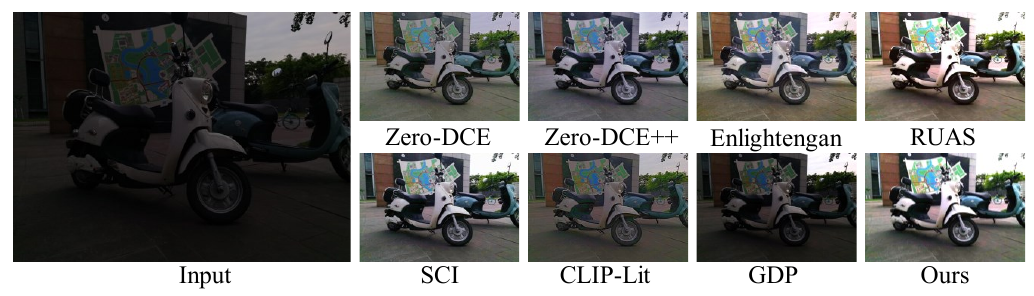}  
         % \vspace{-2em}
        \caption{Visual comparison of low-light enhancement methods on the LSRW dataset.}
        \label{fig:4}
    \end{figure*}

\begin{figure*}[h]
        \centering
        \includegraphics[height=0.4\textwidth,width=\textwidth]{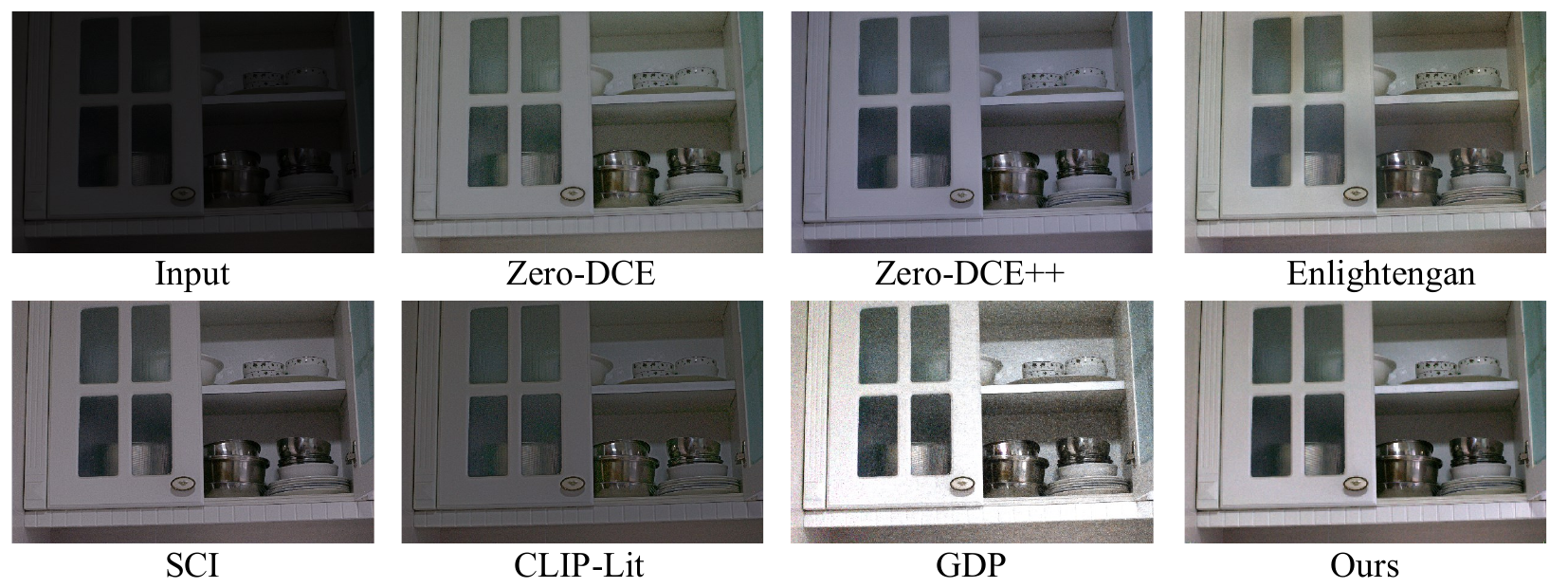}  
         % \vspace{-2em}
        \caption{Visual comparison of low-light enhancement methods on the LOLv1 dataset.}
        \label{fig:5}
    \end{figure*}

\subsection{Comparison with the State-of-the-Art}

{\bf Comparison Methods.} To verify the effectiveness of the method proposed in this paper, we compare it with the state-of-the-art unsupervised learning methods in recent years, i.e., Zero-DCE \cite{guo2020zero}, Zero-DCE++ \cite{li2021learning}, RUAS \cite{liu2021retinex}, Enlightengan \cite{jiang2021enlightengan}, SCI \cite{ma2022toward}, CLIP-Lit \cite{liang2023iterative} and GDP \cite{fei2023generative}.
% \begin{figure*}[h]
%         \centering
%         \includegraphics[height=0.3\textwidth,width=\textwidth]{zfig4.pdf}  
%          \vspace{-2em}
%         \caption{Visual comparison of low-light enhancement methods on the LSRW dataset.}
%         \label{fig:4}
%     \end{figure*}

% \begin{figure*}[h]
%         \centering
%         \includegraphics[height=0.4\textwidth,width=\textwidth]{zfig5.pdf}  
%          \vspace{-2em}
%         \caption{Visual comparison of low-light enhancement methods on the LOLv1 dataset.}
%         \label{fig:5}
%     \end{figure*}

{\bf Quantitative Comparison.} We obtained quantitative results for the other methods using official pre-trained models and running their respective public codes. As shown in Table. \ref{tab:my-table}, our method achieves quantitative performance close to the state-of-the-art on several metrics compared to all compared methods. Among them, we obtain the second-best results in the LOLv1 dataset regarding the full-reference distortion metric PSNR/SSIM. In the LSRW dataset, we obtained the second-best results for the PSNR metric. In addition, for the no-reference metrics NIQE/LOE, our method obtains the lowest NIQE scores on all datasets and the second-best LOE metric evaluation results on both datasets. Our method can better balance the quantitative results of images. This fully demonstrates the superiority of our method and its better generalization ability in real-world scenarios.

{\bf Qualitative Comparison.} For a more intuitive comparison, we report the visual results of all methods in Fig. \ref{fig:4}. By visual comparison, our methods achieve visually pleasing results with improved color and brightness. In contrast, previous state-of-the-art unsupervised learning methods produce artifacts and unnatural tones by producing excessive smoothing or struggling to adapt effectively to degradation factors due to a lack of practical constraints and guidance. For example, Enlightengan \cite{jiang2021enlightengan} produces artifacts, and CLIP-Lit \cite{liang2023iterative} produces excessive color effects. In particular, Diffusion prior-based GDP \cite{fei2023generative} hardly enhances low-light images extensively. Furthermore, we visually compare the LOLv1 test set in Fig. \ref{fig:5}. In contrast, other methods fail to recover enough brightness (SCI \cite{ma2022toward}) or generate too much noise (GDP \cite{fei2023generative}). Our approach provides more substantial constraints, including visually orientated guidance from prompts, thereby producing a more natural visual perception.

\subsection{Ablation Studys} 
\begin{figure*}[t]
        \centering
        \includegraphics[height=0.18\textwidth,width=\textwidth]{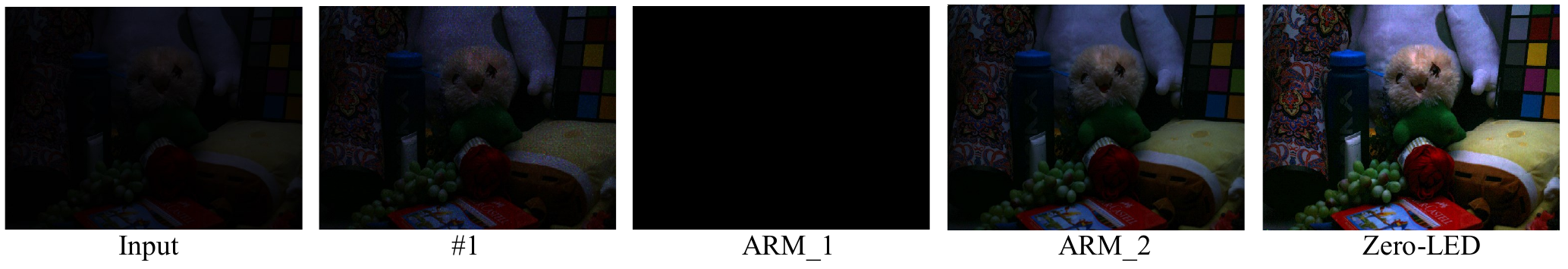}  
         % \vspace{-2em}
        \caption{Qualitative results from ablation studies. Networks with complete structures perform best.}
        \label{fig:6}
    \end{figure*}

% In this subsection, we will perform a series of ablation studies on the initial optimization network and the appearance reconstruction module to explore the most appropriate parameter pairings for the model. We will perform the following judgement analysis in the LSRW test set:
% {\bf The effectiveness of semantic-based and multi-frequency domain appearance reconstruction.}
% The effectiveness of semantic-based and multi-frequency domain appearance reconstruction. We divide the appearance reconstruction module into three versions (i.e., $ARM_1, ARM_2, ARM_3$). In the $ARM_1$ version, we use the Laplace transform and ignore semantic guidance. In the $ARM_2$ version, we use a single frequency domain and semantic guidance. In the $ARM_3$ version, we jointly use complete text-based and multiple-frequency domains for appearance restoration. As shown in the table \ref{tab:6}, it is easy to see that compared to $ARM_1$, the combination of $ARM_2$ semantic and frequency domains produces better enhancement effect. In addition, the test result of $ARM_3$ achieves the optimal result, which proves the effectiveness of the joint guidance of multiple frequency domains.
{\bf The effectiveness of bidirectional optimization training.} To validate the importance of the initial optimization network, we consider two training approaches for ablation studies. We used the LOLv1 dataset for all ablation experiments. "\#1" indicates that the initial optimization network was removed from the overall architecture and trained by directly inputting low-light images for diffusion enhancement. Meanwhile, we used the full Zero-LED for two-way optimization training. Firstly, as shown in Table. \ref{tab:4}, we evaluate the results of \#1, whose enhancement effect is very different from the final results, which illustrates the importance of the initial supervised network. Moreover, as shown in Fig. \ref{fig:6}, we can intuitively conclude that Zero-LED produces clearer results and better perceptual effects than "\#1". This fully validates the iterative optimization effect of our two-way optimization training method.

{\bf The effectiveness of appearance reconstruction based on multi-modal semantics and frequency domain space.} We divide the appearance reconstruction module into three versions by gradually adding module parts. In the $ARM_1$ version, we remove the appearance reconstruction module completely. In the $ARM_2$ version, we add frequency domain spatial reconstruction of images while ignoring multi-modal semantic guidance. Finally, we used the full Zero-LED, added multi-modal semantics, and combined it with the frequency domain space to restore the appearance of the image. As shown in Fig. \ref{fig:6}, in the $ARM_1$ version, we could not reconstruct the image efficiently. With the addition of the frequency domain space, the $ARM_2$ version restores objects in dark areas to a large extent, but the enhanced brightness and colors are still unrealistic. On the other hand, zero LED results in more realistic brightness and optimal perception. This proves the importance of the frequency domain space for image content reconstruction and the effectiveness of multi-modal semantics in guiding image appearance. Furthermore, as shown in Table. \ref{tab:4}, the overall metric evaluation results obtained for $ARM_2$ significantly differed from Zero-LED. We attribute this to the visual friendliness of the text-guided appearance.

% \begin{table}[h]
% \centering
% \renewcommand\arraystretch{1.4}
% \caption{Quantitative evaluation of the enhancement results obtained from different settings. Results using the complete settings are underlined.}
% \begin{tabular}{l|cccc}
% \hline
% Versions  & PSNR↑                         & SSIM↑                        & LPIPS↓                       & FID↓                          \\ \hline
% HFDPM\_v1 & 27.638                        & 0.862                        & 0.215                        & 43.193                        \\ \hline
% HFDPM\_v2 & {\color[HTML]{333333} 28.282} & {\color[HTML]{333333} 0.868} & {\color[HTML]{333333} 0.209} & {\color[HTML]{333333} 41.185} \\ \hline
% HFDPM\_v3 & \textbf{29.212}               & \textbf{0.872}               & \textbf{0.197}               & \textbf{40.987}               \\ \hline
% \end{tabular}
% \label{tab:4}
% \end{table}

\begin{table}[h]
\centering
\renewcommand\arraystretch{1.4}
\begin{tabular}{c|cccc}
\hline
index                         & PSNR↑          & SSIM↑        & NIQE↓        & LOE↓          \\ \hline
\multicolumn{1}{c|}{\#1}      & 12.023      & 0.413       & 12.579       & 132.845       \\ \hline
\multicolumn{1}{c|}{ARM\_1}   & -            & -           & -           & -             \\ \hline
\multicolumn{1}{c|}{ARM\_2}   & 14.559       & 0.512       & 10.613      & 137.836       \\ \hline
\multicolumn{1}{c|}{Zero-LED} & {\ul 16.432} & {\ul 0.562} & {\ul 8.355} & {\ul 148.563} \\ \hline
\end{tabular}
\vspace{10pt}
\caption{Quantitative evaluation of the enhancement results obtained from different settings. Results using the complete settings are {\ul underlined}.}
\label{tab:4}
\end{table}
\vspace{-4em}

\section{Conclusion}
\vspace{-1em}
We propose a bi-directional zero-reference training approach via an initial optimization network and successfully implement a zero-reference trained diffusion model called Zero-LED. First, for the input degraded images, we perform preliminary fitting of the degradation parameters via an initial optimization network and acquire the structural images. We obtain a better lighting estimation with the calibration based on the diffusion model. In addition, we propose a text- and frequency-domain-based appearance reconstruction module for the output restored image, which provides perceptually oriented restoration guidance using a pre-trained visual language model and multiple frequency-domain spaces to guide the restoration of structural content jointly. Metric favorable and perceptually oriented enhancement results are achieved. Experimental results on publicly available benchmark tests show that our approach outperforms competitors in the comprehensive evaluation while providing better stability and generalization.

% \clearpage\mbox{}Page \thepage\ of the manuscript.
% \clearpage\mbox{}Page \thepage\ of the manuscript.
% \clearpage\mbox{}Page \thepage\ of the manuscript.
% \clearpage\mbox{}Page \thepage\ of the manuscript.
% \clearpage\mbox{}Page \thepage\ of the manuscript. This is the last page.
% Now we have reached the maximum length of an ECCV \ECCVyear{} submission (excluding references).
% References should start immediately after the main text, but can continue past p.\ 14 if needed.
% \clearpage  % TODO REVIEW/FINAL: This \clearpage needs to be removed from both review and camera-ready versions.

\bibliographystyle{splncs04}
\bibliography{main}
\end{document}